\documentclass{article}
\usepackage{icais_conference,times}
\usepackage{graphicx}
\usepackage{booktabs}
\usepackage{amsmath,amssymb,bm}
\usepackage{microtype}
\usepackage{booktabs}   
\usepackage{multirow}   
\usepackage{bm}
\usepackage{amssymb}    
\usepackage{amsmath}
\usepackage{pifont}
\usepackage{array}

\newcolumntype{P}[1]{>{\centering\arraybackslash}p{#1}}
\usepackage{makecell}
\usepackage[table]{xcolor}
\usepackage{algorithm}
\usepackage{algorithmic}

\usepackage{hyperref}
\hypersetup{hidelinks}
\usepackage{url}

\title{Simon-SR: Spatially Adaptive Modulation and Visual Prompt Adaptation for Text-Reinforced Super-Resolution}

\author{Haotong Cheng$^*$, Yuxuan Li$^*$, Zijie Cui$^*$\\
College of Electronic Science and Engineering, Jilin University\\
\texttt{\{chenght9923,yxli1923,cuizj1923\}@mails.jlu.edu.cn}}

\iclrfinalcopy

\begin{document}
\maketitle

\begingroup
\renewcommand{\thefootnote}{\fnsymbol{footnote}}
\footnotetext[1]{Equal Contribution.}
\endgroup

\lhead{ICAIS 2026 Extended Abstract Track}

\begin{abstract}
Severe downsampling makes single image super-resolution (SISR) an ill-posed problem, in which the language-guided multi-modal methods are especially vulnerable to erroneous priors and require costly textual annotations. To address these issues, we propose Simon-SR, a multi-modal SISR framework leveraging learnable prompts for efficient semantic mining
and robust text-image fusion. Simon-SR treats textual semantics as learnable latent variables. Specifically, the Contrastive Prompt Learning (CPL)  mines instance-level semantics from unannotated images with frozen CLIP encoders, and Prompt-Guided Spatially Adaptive Refinement (PSAR) injects them through attention-gated multi-modal fusion. On CUB  and COCO2017 at $\times4$ and $\times16$, Simon-SR gains up to 0.50 dB PSNR and 0.0133 SSIM over current SOTA. The project is available at \url{https://github.com/CHT05017/ICAIS26-SimonSR}
\end{abstract}

\section{Motivation and Methodology}
Single Image Super-Resolution (SISR) aims to reconstruct high-quality images from low-resolution inputs, and is an ill-posed problem especially at $\times16$ downsamping. For existing works, pixel-oriented networks often oversmooth fine-grained details~\citep{zhang2018image}. Although the language-guided multi-modal models could recover plausible textures but may be affected by inaccurate language captions~\citep{qu2024xpsr}. To this end, we propose a novel multi-modal SISR
framework termed \textbf{S}patially Adapt\textbf{i}ve \textbf{M}odulation and Visual
Pr\textbf{o}mpt Adaptatio\textbf{n} (Simon-SR), which improves the model's ability to restore fine-grained details while reducing the heavy dependence on textual captions.

Given the groundtruth text-image pairs $\{I_{x}^{HR}, T_{x}^{HR}\}_{x=1}^{N}$, Contrastive Prompt Learning (CPL) associates each instance with $M$ learnable tokens in template  ``A photo of a $[X]_1\ldots[X]_M$.'' Frozen CLIP encoders then produce visual and textual semantics, namely $V_x\  \text{and} \ T_x$~\citep{radford2021learning}. Prompts are optimized by minimizing $\mathcal{L}_{con}=\sum_x\mathcal{L}_{i2t}(x)+\sum_x\mathcal{L}_{t2i}(x)$,
where $\mathcal{L}_{i2t}$ and $\mathcal{L}_{t2i}$ denote the contrastive losses:
\begin{equation}
    \mathcal{L}_{i2t}=-\log \frac{\exp(s(V_x,T_x))}{\sum_{a=1}^B \exp(s(V_x,T_a))},    \mathcal{L}_{t2i}=-\log \frac{\exp(s(V_x,T_x))}{\sum_{a=1}^B \exp(s(V_a,T_x))}
\end{equation}

Prompt-Guided Spatially Adaptive Refinement (PSAR) fuses the semantic anchors with visual features in Progressive Text-Aware Refinement Blocks (PTRBlocks). To obtain the modulated feature $\hat{\bm{f}}_{img}^{(l)}$, we project $\bm{f}_{img}^{(l)}$ into query space through $\mathbf{Q}=\text{Conv}_{1\times1}(\bm{f}_{img}^{(l)})$, and $\bm{f}_{txt}^{(l)}$ into key space through $\mathbf{K}=\text{Linear}(\bm{f}_{txt}^{(l)})$. After reshaping, the spatial attention is computed as $
    \mathcal{A}=\sigma(\mathbf{Q}\cdot\mathbf{K}^{\top} / ({\sqrt{C/r}}))
$. $\mathcal{A}$ encodes the spatial correlation between visual content and textual counterparts. We then define the affine parameters as $\Delta \bm{\Gamma}=P_{\gamma}(\bm{f}_{img}^{(l)})$ and $\Delta \bm{B}=P_{\beta}(\bm{f}_{txt}^{(l)})$, where $P_{\gamma}(\cdot)$ and $P_{\beta}(\cdot)$ denote two linear projections. The final transformation is formulated as:
\begin{equation}
\widehat{\bm f}_{\mathrm{img}}^{(l)}=
(\bm 1+\Delta\bm\Gamma\odot\mathcal{A})\odot\bm f_{\mathrm{img}}^{(l)}
+\Delta\bm B\odot\mathcal{A}.
\label{eq:psar}
\end{equation}
Low-relevance regions begin as identity mappings, and regions of interest are selectively enhanced. Training uses $\mathcal{L}_{\mathrm{sr}}+\mathcal{L}_{\mathrm{percep}}+0.02\mathcal{L}_{\mathrm{adv}}$, namely the combination of reconstruction loss, perceptual loss, and text-conditioned adversarial losses.

\begin{figure}[t]
\centering
\includegraphics[width=\linewidth,trim=0.14cm 3.81cm 0.20cm 2.60cm,clip]{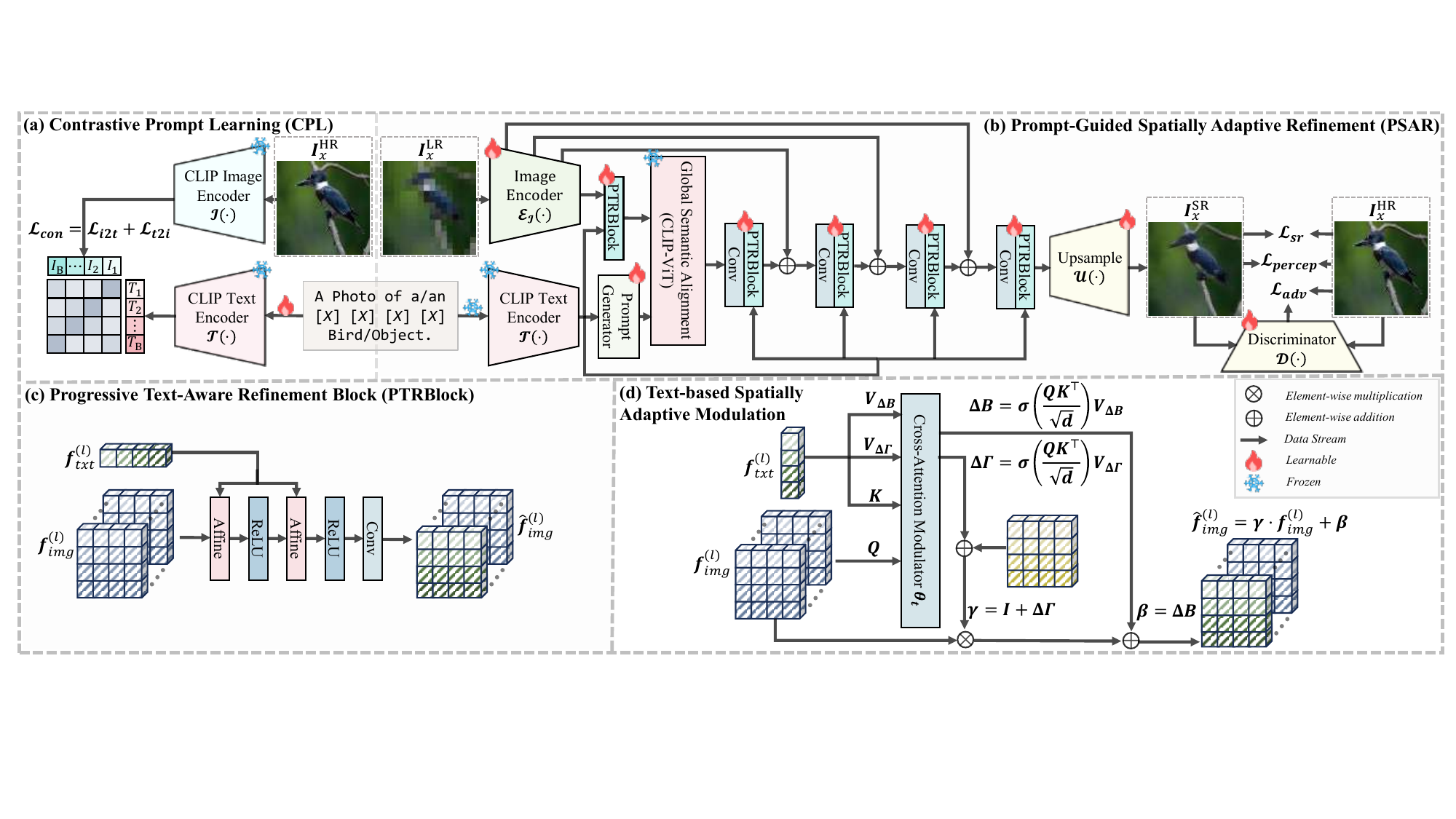}
\vspace{-0.4cm}
\caption{Simon-SR learns semantic prompts with CPL and performs vision-language fusion with PSAR, which includes PTRBlocks and spatially adaptive modulation.}
\vspace{-0.3cm}
\end{figure}

\begin{table}[!t]
\centering
\caption{Representative $\times 4$ and $\times 16$ results against the strongest SOTA baseline.}
\label{tab:results}
\setlength{\tabcolsep}{6pt}
\begin{tabular}{clcccccc}
\toprule
\multirow{2}{*}{Scale}
& \multirow{2}{*}{Models}
& \multicolumn{3}{c}{CUB}
& \multicolumn{3}{c}{COCO2017} \\
\cmidrule(lr){3-5}
\cmidrule(lr){6-8}
& & PSNR$\uparrow$ & SSIM$\uparrow$ & LPIPS$\downarrow$
& PSNR$\uparrow$ & SSIM$\uparrow$ & LPIPS$\downarrow$ \\
\midrule

& Baseline~\cite{hu2025clip}
& 28.44 & 0.8409 & 0.0996
& 26.23 & 0.7607 & \textbf{0.1185} \\
\multirow{-2}{*}{$\times 4$}
& \textbf{Simon-SR}
& \textbf{28.53} & \textbf{0.8452} & \textbf{0.0977}
& \textbf{26.38} & \textbf{0.7687} & 0.1220 \\
\midrule

& Baseline~\cite{hu2025clip}
& 21.41 & 0.5486 & 0.3184
& 20.10 & 0.4414 & 0.3840 \\
\multirow{-2}{*}{$\times 16$}
& \textbf{Simon-SR}
& \textbf{21.91} & \textbf{0.5615} & \textbf{0.3086}
& \textbf{20.15} & \textbf{0.4547} & \textbf{0.3804} \\
\bottomrule
\end{tabular}
\vspace{-0.3cm}
\end{table}

\section{Experiments and Findings}
We evaluate Simon-SR against the strongest SOTA baseline ~\citep{hu2025clip} on CUB~\citep{wah2011caltech} and COCO2017~\citep{lin2014microsoft} at $\times4$ and $\times16$. Table~\ref{tab:results} shows that Simon-SR consistently outperforms current SOTA in both visual quality and structual fidelity. Particularly, Simon-SR outperforms SOTA by 0.50 dB PSNR in $\times16$ CUB, validating the effectiveness of the proposed framework. To conclude, Simon-SR thus provides an efficient method for annotation-free semantic mining and robust vision-language fusion through spatially adaptive modulation.

\subsubsection*{AI-Use Disclosure and Conflict of interest}
Generative AI assisted in language polishing, and the authors have carefully checked the generated contents. Manuscript drafting, reported experiments, and codes are conducted manually without any use of AI. The authors declare that they have no conflict of interest.

{\small
\setlength{\bibsep}{0pt}
\renewcommand{\bibfont}{\fontsize{8}{9}\selectfont}
\bibliography{references}
\bibliographystyle{icais_conference}
}

\end{document}